\documentclass[12pt, reqno]{amsart}
\usepackage{amsmath, amsthm, amscd, amsfonts, amssymb, graphicx, mathtools, dsfont, pifont}
\usepackage[table,xcdraw]{xcolor}
\usepackage{helvet, quiver, subcaption}
\usepackage[bookmarksnumbered, colorlinks, plainpages]{hyperref}
\usepackage{diagbox}
\usepackage[flushmargin]{footmisc}
\usepackage{enumitem}
\usepackage{multirow}

\newtheorem{theorem}{Theorem}
\newtheorem{lemma}[theorem]{Lemma}
\newtheorem{proposition}[theorem]{Proposition}
\newtheorem{corollary}[theorem]{Corollary}
\newtheorem{fact}[theorem]{Fact}
\theoremstyle{definition}
\newtheorem{definition}[theorem]{Definition}
\newtheorem{example}{Example}

\newtheorem{question}[theorem]{Question}
\theoremstyle{remark}
\newtheorem{remark}[theorem]{Remark}
\numberwithin{equation}{section}

\newtheorem*{lemman}{Lemma}
\newtheorem*{theoremn}{Theorem}

\setlist[itemize]{topsep=0.3em,
  itemsep=0.2em,
  parsep=0pt}
  
\setlist[enumerate]{topsep=0.3em,
  itemsep=0.2em,
  parsep=0pt}

\newcommand{\N}{\mathbb{N}}
\newcommand{\R}{\mathbb{R}}

\newcommand{\Q}{\mathbb{Q}}

\newcommand{\A}{\mathcal{A}}

\renewcommand{\H}{\mathcal{H}}
\newcommand{\Db}{\mathcal{D}}
\newcommand{\F}{\mathcal{F}}
\newcommand{\G}{\mathcal{G}}
\newcommand{\X}{\mathcal{X}}
\newcommand{\Y}{\mathcal{Y}}
\renewcommand{\S}{\mathcal{S}}
\newcommand{\D}{\mathbb{D}}

\newcommand{\eps}{\varepsilon}
\renewcommand{\phi}{\varphi}

\DeclareMathOperator{\dom}{dom}

\DeclareMathOperator{\code}{code}

\DeclareMathOperator{\im}{im}
\newcommand{\res}{\mathord{\upharpoonright}}

\newcommand{\summe}[2]{\sum\limits_{#1}^{#2}}

\DeclareMathOperator*{\argmin}{arg\,min}

\DeclareMathOperator{\supp}{supp}
\newcommand{\seq}{\subseteq}
\newcommand{\dash}{\dashrightarrow}
\renewcommand{\setminus}{\backslash}

\renewcommand{\emptyset}{\varnothing}
\newcommand{\prob}[2]{\underset{#1}{\mathrm{Prob}}\left[#2\right]}

\newcommand{\infi}[1]{\inf\limits_{#1}}
\newcommand{\mini}[1]{\min\limits_{#1}}
\newcommand{\maxi}[1]{\maxi\limits_{#1}}
\DeclareMathOperator{\vc}{VCdim}
\DeclareMathOperator{\evc}{eVCdim}

\renewcommand{\u}[1]{\underline{#1}}

\newcommand{\no}{\textcolor{red}{\ding{56}}}
\newcommand{\yes}{\textcolor{green}{\ding{52}}}
\newcommand{\?}{\textcolor{blue}{\textbf{?}}}

\DeclareMathOperator{\Hfin}{\mathcal{H}_{\mathrm{fin}}}
\DeclareMathOperator{\hfin}{\mathcal{H}_{\mathrm{fin}}}

\DeclareMathOperator{\mnu}{m^{NU}_\mathcal{H}}

\newcommand{\one}{\mathds{1}}

\title[RER Classes and the Fundamental Theorem]{Recursively Enumerably Representable Classes and Computable Versions of the Fundamental Theorem of Statistical Learning}

\author[D.~Kattermann]{David Kattermann$^{1,3}$}
\thanks{\noindent$^{1}$\tiny Institute for Symbolic Artificial Intelligence, Johannes Kepler University Linz, Austria}

\author[L.~S.~Krapp]{Lothar Sebastian Krapp$^{2,3}$}
\thanks{\noindent$^{2}$\tiny Institute for the Interdisciplinary Study of Language Evolution, University of Zurich, Switzerland}
\thanks{\noindent$^{3}$\tiny Department of Mathematics and Statistics, University of Konstanz, Germany}

\begin{document}
\setcounter{page}{1}

\color{darkgray}{
\noindent 

\centerline{}

\centerline{}

\begin{abstract}%
We study computable probably approximately correct (CPAC) learning, where learners are required to be computable functions. It had been previously observed that the  Fundamental Theorem of Statistical Learning, which characterizes PAC learnability by finiteness of the Vapnik--Chervonenkis (VC-)dimension, no longer holds in this framework. Recent works recovered analogs of the Fundamental Theorem in the computable setting, for instance by introducing an effective VC-dimension. In this work, we investigate the relationship between CPAC learning and recursively enumerable representable (RER) classes, hypothesis classes whose members can be algorithmically listed, in the context of the Fundamental Theorem. We demonstrate that the RER property is deeply connected to CPAC learning by characterizing several notions of CPAC learnability via the existence of certain RER classes realizing the same samples. We further establish that the RER property alone is sufficient to guarantee nonuniform CPAC learnability and give a sufficient condition for CPAC learnable classes to be RER. Other results show that the effective VC-dimension can take arbitrary values above the traditional one and we note that the two dimensions coincide given the existence of a computable empirical risk minimizer. This recovers classical PAC bounds for most practically relevant classes and establishes a family of examples separating several notions of learnability.
\end{abstract}
\maketitle


\section{Introduction}
\textit{Probably Approximately Correct} (PAC) learning is a well-known framework in statistical learning theory to describe how machines are able to generalize in supervised classification tasks. The theory was developed in the second half of the 20th century. More recently, \cite{agar} initiated a study of \textit{computable} PAC (CPAC) learning, where the learning process is required to be effective in the sense of theoretical computer science. This forms an intermediate setup between the abstract study of PAC learning, called \textit{VC-theory} after~\cite{vc}, and the framework from \cite{valiant}, where learning processes are even required to be efficient. The restriction to computable learners can be seen as a minimal requirement to model automated learning. It turns out that this requirement has drastic effects on classical results and therefore received a considerable amount of attention in recent years, see~\cite{sterk, online, gourdeau, gourdeau2, rose, rose2, hta}.

A hypothesis class $\H$ is PAC learnable if there exists a function $\A$, called \emph{learner}, that, with arbitrarily high probability, outputs a hypothesis with arbitrarily low generalization error, given a sufficient number of training examples. Several variations of this definition have been considered before (see Table~\ref{table pac}) and the literature is not entirely consistent when referring to PAC learning. For example, some authors require the existence of a \emph{proper} learner, which maps into $\H$, while others allow arbitrary (or \emph{improper}) learners. Such inconsistencies are neglectable in the traditional framework, due to a powerful result, known as the \textit{Fundamental Theorem of Statistical Learning} by~\cite{blumer}. It states that (under mild measurability assumptions) PAC learnability in all its variants is equivalent to finiteness of a purely combinatorial measure of $\H$, the \textit{Vapnik--Chervonenkis dimension} (VC-dimension).

The central motivation for this paper is to achieve a better understanding of how the Fundamental Theorem of Statistical Learning behaves in the CPAC learning framework. A key insight of \cite{agar} was that the Fundamental Theorem of Statistical Learning generally fails once learners are required to be computable functions. In particular, the characterizations of the various notions of PAC learnability via finiteness of the VC-dimension are lost, which leads to a diverse landscape of intermediate notions. Previous works already recovered parts of the Fundamental Theorem by characterizing some intermediate notions of CPAC learnability in analogy to the classical theorem. Most importantly, \cite{rose} introduced a computable analog of the VC-dimension, the \textit{effective} VC-dimension, whose finiteness characterizes CPAC learnability in the (improper) agnostic setting, where arbitrary sample distributions are considered.

Alongside CPAC learnability as the computability analog of PAC learnability, \cite{agar} proposed \textit{recursively enumerable representability} (RER) as a computability analog for hypothesis classes. The RER property states that the members of a hypothesis class can be listed by an algorithm. It is known that for RER classes CPAC learnability in the realizable setting, where the sample distributions are assumed to be compatible with the considered hypothesis class, is characterized solely by finiteness of the VC-dimension. This already indicates that the RER property is closely connected to CPAC learning. However, previous works have treated the RER property rather as a convenient technical condition.

In this paper we show that the RER property is inherently connected to CPAC learnability. Our results show that it is essentially enough to consider RER classes for most types of CPAC learnability, in the sense that an arbitrary class is only learnable, if it is sufficiently similar to some learnable RER class. Hence, working under the RER assumption is not really a restriction. This aligns with recent work by \cite{hta}, where it is argued that issues with computability in PAC learning are ``unnatural'' from a practical point of view, as previously considered problematic classes were defined using some numbering of programs. The authors demonstrated that under mild regularity assumptions, such as being RER, any ``natural'' PAC learnable class is already CPAC learnable in the strongest sense. 
In Section \ref{sec evc} we complement these findings by showing that, while in general the effective VC-dimension can take arbitrary values above its traditional counterpart, the two dimensions coincide for any class that is CPAC learnable in the strongest sense.

\subsection{Organization and contributions of this paper.}
Our overall approach is to analyze the interplay between RER classes, CPAC learning and the effective VC-dimension. In Section~\ref{sec prelims} we give a detailed overview of key notions and results from (C)PAC learning to make some connections between recent results and the Fundamental Theorem more explicit. After that, we begin in Section~\ref{chap rer} by investigating how the RER property relates to CPAC learnability by looking at the realized samples of a hypothesis class. This perspective offers a whole plethora of novel necessary and sufficient conditions for different types of learnability. The theme of these results is always to characterize a specific notion of CPAC learnability of an arbitrary hypothesis class $\H$, solely via the existence of some RER class $\G$ that satisfies some inclusion relation between $\G$ and $\H$, as well as between the sets of realized samples $\S_\G$ and $\S_\H$, and some additional property ensuring learnability of $\G$. These results are summarized in Table~\ref{table hg}.
Proposition \ref{prop uip} provides a partial converse by showing that properly, realizably CPAC learnable classes, in which each hypothesis is identifiable from a sample (which includes classes of finite teaching dimension), are necessarily RER.

Section~\ref{sec nupac} studies the relaxed notion of \textit{nonuniform} PAC learning in the computable setting, also introduced in \cite{agar}. In Theorem \ref{thm nucpac} we adapt structural risk minimization to show that nonuniform CPAC learnability can be guaranteed for any RER class, regardless of its VC-dimension. Since this nonuniform learnability is even possible with a proper learner, this result transfers to all subclasses of RER classes and some extensions.

In Section~\ref{sec evc} we compare the classical and effective VC-dimension. On the negative side, Theorem~\ref{thm evc} generalizes a construction from \cite{rose} to show that the effective VC-dimension can take arbitrary values above the traditional one, even for RER classes. On positive side, Theorem~\ref{prop equal} identifies an important boundary case, namely, under the assumption of the existence of a total computable ERM the two measures coincide. This recovers quantitative PAC bounds in the CPAC framework, something that cannot be guaranteed in general by our negative result (Remark \ref{rem bound}). As a byproduct of Theorem~\ref{thm evc}, we also get a whole family of new examples separating various notions of learnability (Remark~\ref{rem examples}). We conclude in Section~\ref{sec sum} with an overview of our findings and an open question. Appendix~\ref{sec answer} contains answers to two questions posed in the published version of this paper.

\begin{table}[ht]
\centering
\resizebox{\textwidth}{!}{
\begin{tabular}{|c||c||ccc||c|}
\hline
\multirow{2}{*}{Learnability of $\H$} & \multirow{2}{*}{} & \multicolumn{3}{c||}{$\exists$ RER class $\G$} & Result \\ \cline{3-5}
 &  & \multicolumn{1}{c|}{Inclusion}  & \multicolumn{1}{c|}{Samples} & \multicolumn{1}{c||}{Property of $\G$} & number \\ \hline
proper realizable SCPAC & $\Leftrightarrow$ & \multicolumn{1}{c|}{$\G\seq \H$} & \multicolumn{1}{c|}{$\S_\H=\S_\G$} & \multicolumn{1}{c||}{$\vc(\G)<\infty$}  & \ref{cor rer}\\ \hline
proper agnostic SCPAC & $\Leftrightarrow$ & \multicolumn{1}{c|}{$\G\seq \H$} & \multicolumn{1}{c|}{$\S_\H=\S_\G$} & \multicolumn{1}{c||}{$\vc(\G)<\infty + \text{comp. ERM}$} & \ref{thm rer scpac} \\ \hline
agnostic CPAC & $\Leftrightarrow$ & \multicolumn{1}{c|}{$\G\seq \Hfin$} & \multicolumn{1}{c|}{$\S_\H\seq \S_\G$}  & \multicolumn{1}{c||}{$\evc(\G)<\infty$} & \ref{cor agn} \\ \hline
proper nonuniform CPAC & $\Rightarrow$ & \multicolumn{1}{c|}{$\G\seq \H$} & \multicolumn{1}{c|}{$\S_\H=\S_\G$} & \multicolumn{1}{c||}{-} & \ref{cor nu} \\ \hline
nonuniform CPAC & $\Leftarrow$ & \multicolumn{1}{c|}{$\G\supseteq \H$} & \multicolumn{1}{c|}{$\S_\H\seq\S_\G$} & \multicolumn{1}{c||}{-}  & \ref{cor cont} \\ \hline
\end{tabular}
}
\captionsetup{width=\textwidth, skip=7pt}
\caption{Necessary and sufficient conditions for various notions of learnability for an arbitrary hypothesis class $\H$ via the existence of an RER class $\G$ satisfying three conditions.}\label{table hg}
\end{table}

\section{Preliminaries}\label{sec prelims}
\noindent Throughout this work, $\N$ denotes the set of natural numbers including $0$ and $\N_+\coloneqq\N\setminus \{0\}.$  For $n\in \N_{+}$ we define $[n]\coloneqq \{1,\hdots, n\}.$ Let $X$ and $Y$ be sets. The cardinality of $X$ is denoted by $|X|.$  We write $f\colon X\dash Y$ to express that $f$ is a \textit{partial} function, meaning that its domain $\dom(f)$ may be a proper subset of $X.$ If $\dom(f)=X$ we say that $f$ is \textit{total} and indicate this by writing $f\colon X\to Y.$ The set of all total functions from $X$ to $Y$ is denoted by $Y^X.$ If $0\in Y$, then the \textit{support} of $f\colon X\dash Y$ is $\supp(f)\coloneqq f^{-1}(Y\setminus\{0\}).$

In what follows, we work with some \textit{domain} $\X$ and fix the \textit{label set} $\Y\coloneqq \{0,1\}.$  In general, $\X$ can be the underlying set of an arbitrary measure space. But since we are only interested in \textit{computable} PAC learning, \textbf{throughout this paper we assume $\X=\N$} or any countable set that can be computably encoded into the natural numbers. We also fix the powerset as $\sigma$-algebra on $\X\times \Y.$ A \textit{hypothesis} is any (measurable) function $h\colon \X\to \Y,$ and a \textit{hypothesis class} $\H$ is any nonempty set of hypotheses. A \textit{sample} is a finite tuple of labeled examples, i.e.~$S=((x_i,y_i))_{i\in [m]}$ with $(x_i,y_i)\in \X\times \Y.$ We write $|S|$ for the length of a sample and  $\S\coloneqq \bigcup_{m\in \N_+}(\X\times \Y)^m$ for the set of all samples. A \textit{learner} is any (partial) function $\A\colon \S\dash \Y^\X$ outputting only hypotheses. Furthermore, we write $\Db^*$ for the set of all distributions on our $\sigma$-algebra on $\X\times \Y.$  For $\D\in \Db^*$ we define the \textit{loss} of a hypothesis $h$ as $L_\D(h)\coloneqq \D\big(\{(x,y)\colon h(x)\neq y\}\big)$. For a sample $S=((x_i,y_i))_{i\in [m]}$ the \textit{empirical loss} of $h$ is defined as $L_S(h)\coloneqq \frac{1}{m}|\{i\in [m]\colon h(x_i)\neq y_i\}|.$ Following the notation in \cite{sbd}, we write 
$\prob{S\sim \D^m}{\cdots}$ for the probability $\D^m(\{S\colon \cdots\}),$ where $\D^m=\bigotimes_{i=1}^m\D.$ Given a hypothesis class $\H\seq\Y^\X$ we write $\S_\H\coloneqq\big\{((x_i,h(x_i)))_{i\in [m]}\colon m\in \N_+, x_i\in \X, h\in \H\big\}$ for the set of all \textit{realized} samples of $\H.$ Similarly, we write $\Db_\H\coloneqq \{\D\in\Db^*\colon \exists h\in \H\colon L_\D(h)=0\}.$
By convention we set $L_\D(\A(S))=1$ if $S\notin \dom(\A).$
An overview of all used acronyms can be found at the end of this section.

\begin{definition}
Let $\H\seq \Y^\X$ be a hypothesis class, $\Db\seq \Db^*$,  $\A\colon\S\dash \Y^\X$ be a learner and $m_\H\colon \N^2\to \N.$
Then $\H$ is said to be \textbf{PAC learnable} with respect to $\mathbb{D}$ with learner $\A$ and sample size function $m_\H\colon \N^2\to \N$, if for all $a, b\in \N_+$ and every distribution $\D\in \mathcal{D}$ it holds for every $m\geq m_\H(a, b)$ that
\begin{equation*}\label{pac def}
		\prob{S\sim \D^m}{L_\D(\A(S))\leq \infi{h\in \H}L_\D(h)+1/a}\geq 1-1/b.
\end{equation*}
\end{definition}

\noindent The objects $\Db, \A$ and $m_\H$ are usually omitted unless we want to point them out explicitly. By imposing conditions on distributions, learners and sample size functions, we introduce some variations of this definitions, also summarized in Table~\ref{table pac}.
\begin{definition}\label{def notions}
Let $\H$ be a hypothesis class.
\begin{enumerate}
\item $\H$ is said to be \textbf{agnostically} PAC learnable if it is PAC learnable w.r.t.~$\Db^*$ and with some total learner $\A.$
\item $\H$ is said to be \textbf{realizably} PAC learnable if it is PAC learnable w.r.t.~$\Db_\H$ and with some learner $\A$ satisfying $\S_\H\seq \dom(\A).$
\item $\H$ is said to be \textbf{properly} PAC learnable if it is PAC learnable with some learner $\A$ satisfying $\im(\A)\seq \H.$
\item $\H$ is said to be \textbf{strongly} PAC learnable if it is PAC learnable with some computable sample size function $m_\H$.
\end{enumerate}
\end{definition}

\noindent The term \textbf{improper} is used by some authors to stress a \textit{potential} non-proper learnability, but does not exclude proper learnability. Following \cite{rose}, we do not use this term and instead only mention proper learnability, when it is guaranteed. The notion of strong PAC learning is due to \cite[Definition 6]{sterk}.

\begin{table}[ht]
\centering
\resizebox{\textwidth}{!}{
\begin{tabular}{|l||*{3}{c|}}\hline
\backslashbox{Notions of PAC}{Conditions on}
&\begin{tabular}[c]{@{}c@{}}Set of\\ distributions $\Db$\end{tabular} &Learner $\A$ & \begin{tabular}[c]{@{}c@{}}Sample size\\ function $m_\H$\end{tabular}\\\hline\hline
Agnostic &$\Db=\Db^*$&$\dom(\A)=\S$&-\\\hline
Realizable &$\Db=\Db_\H$&$\S_\H\seq \dom(\A)$&-\\\hline
Proper &-&$\im(\A)\seq \H$&-\\\hline
Improper &-&-&-\\\hline
Strong &-&-&computable\\\hline
Computable  &-&computable&-\\\hline
\end{tabular}
}
\caption{Variations of the definition of PAC learning.}\label{table pac}
\end{table}

The below famous theorem due to~\cite{blumer} shows that in the traditional framework all notions of PAC learning are equivalent. Our formulation is a modern one that is adapted from \cite[Section 6.4]{sbd}.
\begin{definition}\label{def vc}
Let $\H\seq\Y^\X$ be a hypothesis class and $k\in \N.$ 
A \textbf{$k$-witness} is a function $w\colon \X^{k+1}\to \Y^{k+1}$ such that for any $x_0,\hdots, x_k\in \X$ it holds $w(x_0,\hdots, x_k)\neq (h(x_0),\hdots, h(x_k))$ for any $h\in \H.$ The \textbf{VC-dimension} of $\H$, $\vc(\H)$,  is defined as the smallest natural number $k$ for which a $k$-witness for $\H$ exists and $\vc(\H)=\infty$ if no $k$-witnesses exist for $\H$.
\end{definition}

\begin{definition}\label{def erm}
Let $\H$ be a hypothesis class. A learner $\A$ is called an \textbf{empirical risk minimizer} (ERM) for $\H$ if $\A(S)\in \argmin_{h\in \H}L_S(h)$ for every $S\in \dom(\A).$ 
\end{definition}

\begin{theorem}[Fundamental Theorem of Statistical Learning]\ 

\noindent For any\footnote{A formally correct statement for  general domains $\X$ needs an additional measure-theoretic assumption called ``well-behavedness'' of $\H$. We do not need this here, since we only work on countable domains. An extensive discussion of measurability aspects in the Fundamental Theorem can be found in \cite{laura}.} hypothesis class $\H$ the following are equivalent:
\begin{enumerate}
\item $\H$ is agnostically PAC learnable.
\item $\H$ is realizably PAC learnable.
\item $\H$ is properly PAC learnable (with any ERM).
\item $\H$ is strongly PAC learnable.
\item $\vc(\H)<\infty$.
\end{enumerate}
\end{theorem}

Unfortunately, the Fundamental Theorem does not hold in the following computable framework for PAC learning, introduced in \cite[Definition 8]{agar}. We do not assume any background in computability theory here beyond the basic notion of computability, for instance via Turing machines \cite{turing}. We refer the unfamiliar reader to standard textbooks such as \cite{weihrauchcomp, soare}. In the following, an \textit{encoding} of a set refers to its image under an injective mapping to $\N$. Most importantly, any set of tuples, like $\S,$ can be encoded using goedelization via prime numbers $(x_i)_{i\in [m]}\mapsto \prod_{i=1}^mp_i^{x_i+1},$ where $p_i$ is the $i$-th prime number.
Similarly, a \textit{code} is a natural number uniquely assigned to each program in any model of computability that allows executing the program on some universal machine (see \cite[p.\,10]{soare}).

\begin{definition}\label{def cpac}
A function $\A\colon \S\dash \Y^\X$ is called a \textbf{computable learner}\index{Learner!computable} if there exists a computable function $\A'\colon \N \dash \N$ of $\A$ whose domain is the encoding of $\dom(\A)$ and such that for all $S\in \dom(\A)$ the function $\A'$ outputs the code of a program computing $\A(S).$ A hypothesis class is said to be \textbf{computably PAC learnable} (CPAC) if it is PAC learnable with respect to a computable learner. We abbreviate \textit{strong} CPAC learnability as \textbf{SCPAC}.
\end{definition}

Note that according to this definition the learner $\A$ itself is not necessarily a computable function, but rather can be identified with some computable realization $\A'.$ In practice, this distinction is purely formal, since outputs can be uniquely recovered from their codes. For clarity, and in line with common practice, we will therefore work directly with $\A.$

\begin{remark}\label{rem uncomp}
As noted in \cite[Remark 4]{rose}, the definition of CPAC learnability does not require the class $\H$ to only contain computable hypotheses, but only the learner to output computable ones. An example of a (properly) CPAC learnable class containing uncomputable hypotheses is given in Appendix~\ref{sec answer}. 
Also note that the CPAC framework can be generalized to certain continuous domains, instead of just countable ones, using the framework of computable analysis, see \cite{ack}. While some results from the countable setting carry over, this is, however, not a strong generalization as discussed in \cite[Appendix B]{david}.
\end{remark}

In the following we state the known analogs of the Fundamental Theorem of Statistical Learning for different variants of CPAC learning. The first result was originally only formulated in the agnostic setting, but the proof given in the original source can be modified to also work in the realizable setting by simply adapting the domain of the involved learners, see \cite[p.\,65ff]{david}.

\begin{fact}[Fundamental Theorem for proper SCPAC, \protect{\cite[Theorem 8]{sterk}}]\label{fact scpac}\

\noindent A hypothesis class $\H$ is properly agnostically SCPAC learnable if and only if $\vc(\H)<\infty$ and there exists some total computable ERM for $\H.$ Similarly, $\H$ is properly, realizably SCPAC learnable if and only if $\vc(\H)<\infty$ and there exists a computable ERM $\A$ for $\H$ with $\S_\H\seq\dom(\A).$
\end{fact}

In search of a replacement for the VC-dimension, \cite[Definition 9]{rose} proposed the following measure.
\begin{definition}\label{def evc}
The \textbf{effective VC-dimension} of a hypothesis class $\H$ is
\[\evc(\H)\coloneqq \inf\{k\in \N\colon \text{there is a computable $k$-witness for $\H$}\}.\]
\end{definition}

\begin{fact}[Fundamental Theorem for agnostic CPAC, \protect{\cite[Theorem 11]{rose}}]\label{fact evc rose}
\noindent For any hypothesis class $\H$ the following are equivalent:
\begin{enumerate}
    \item $\H$ is agnostically CPAC learnable.
    \item $\H$ is agnostically SCPAC learnable.
    \item There is some properly agnostically SCPAC learnable class $\H'\supseteq \H.$
    \item $\evc(\H)<\infty.$
\end{enumerate}
\end{fact}

Alongside computable PAC learning, Agarwal et al.~\cite[Definition 6]{agar} also introduced the following computability notion for hypothesis classes. The terminology is analogous to sets $M\seq \N^k$, which are called \textit{recursively enumerable} if there exists an algorithm that lists exactly the elements of $M$, or equivalently $M$ is the domain or range of a computable function.

\begin{definition}\label{def rer}
A hypothesis class $\H$ is said to be \textbf{recursively enumerable representable} (RER) if there exists an algorithm listing all codes of programs encoding the hypotheses from $\H.$
\end{definition}

In other words, a hypothesis class is RER if it can be represented by a recursively enumerable set of codes. Note that this forces RER classes to only contain computable functions. Agarwal et al.~\cite[Definition 5]{agar} also introduced a stronger notion of \textit{decidably representable} (DR) classes. This is not discussed here as it does not provide additional insights.

\begin{example}[List representation, \protect{\cite[Remark 7]{agar}}]\label{ex rer}
A very important RER class is $\Hfin\coloneqq \{h\in \Y^\X\colon |\supp(h)|<\infty\}$, the class of all finitely-supported hypotheses. 
Each member of this class can be uniquely represented by its support. Such a list representation not only makes $\Hfin$ RER, but also enables one to computably access the supports from a listing. 
An important RER subclass of $\Hfin$ is the class $\H^{(k)}$ of all hypotheses with support size at most $k$ for $k\in \N.$ It holds $\vc(\H^{(k)})=\evc(\H^{(k)})=k,$ as the function $\u{x}\mapsto (1,\hdots, 1)$ is a $k$-witness for $\H^{(k)}.$
\end{example}

\begin{fact}[Fundamental Theorem for RER classes, \protect{\cite[Theorem 10]{agar}}]\label{fact real rer} \

\noindent For any RER class $\H$ the following are equivalent:
\begin{enumerate}
\item $\H$ is realizably CPAC learnable.
\item $\H$ is properly, realizably SCPAC learnable.
\item $\vc(\H)<\infty.$
\end{enumerate}
\end{fact}

\section{Recursively Enumerably Representable Classes}\label{chap rer}
\noindent In this section we investigate the interplay between RER classes and CPAC learnability by looking at the realized samples $\S_\H=\{S\in \S\colon \exists h\in \H\colon L_S(h)=0\}$ of a hypothesis class $\H$. Recall from Fact \ref{fact real rer} that any RER class with finite VC-dimension is realizably CPAC learnable. The converse is not true in general as observed in \cite[Example 3]{sterk}. The example given therein extends an RER class to a non-RER one, in a way such that no new samples get realized. This maintains the VC-dimension and the existence of an ERM and therefore the proper SCPAC learnability. The following theorem shows that every non-RER example for proper SCPAC learnability is necessarily of that form. This result, like the later ones of similar flavor, is based on the following consequence of the definition of computable learners.
\begin{lemma}\label{lem image}
The image of a computable learner is an RER class.
\end{lemma}
\begin{proof}
Let $\A'\colon \N\dash \N$ be a computable realization of some learner $\A$.  Since $\A'$ is computable, its domain, which is an encoding of $\dom(\A),$ can be computably listed. So we get a listing of codes of programs computing the hypotheses from the image of $\A$ by successively inputting a listing of $\dom(\A')$ into $\A'.$
\end{proof}

\begin{theorem}\label{thm rer scpac}
A hypothesis class $\H$ is properly agnostically SCPAC learnable if and only if it contains some RER class $\G \seq  \H$ that is properly agnostically SCPAC learnable and satisfies $\S_\H = \S_\G$. The same statement holds with ``realizably'' instead of ``agnostically''.
\end{theorem}
\begin{proof} We only consider the agnostic case, the realizable one is analogous.

\noindent `$\Rightarrow$' By Fact \ref{fact scpac} there exists a computable ERM $\A$ for $\H$, and $\G\coloneqq \im(\A)$ is an RER subclass of $\H$ by Lemma \ref{lem image} with $\vc(\G)\leq \vc(\H)<\infty.$
Such a learner $\A$ is also a computable ERM for $\G$. Indeed, $\G\seq \H$ implies $\S_\G\seq \S_\H$ and for any $S\in \dom(\A)$ it holds
\[\A(S)\in \G\cap\argmin\limits_{h\in \H}L_S(h)\seq \argmin\limits_{g\in \G}L_S(g). \]
So the class $\G$ is properly SCPAC learnable by Fact \ref{fact scpac}.
Moreover, for any $S\in \S_\H\seq \dom(\A)$ it holds $L_S(\A(S))=\min_{h\in \H}L_S(h)=0.$
Due to $\A(S)\in \G,$ this shows $S\in \S_\G$ and thus $\S_\H=\S_\G.$

\noindent `$\Leftarrow$' By Fact \ref{fact scpac} it holds $\vc(\G)<\infty$ and there exists some computable ERM for $\G$.
The assumption $\S_\H=\S_\G$ implies $\vc(\H)=\vc(\G)<\infty$. Moreover, it yields that for any $h\in \H$ and any $S\in \S$ there is some $g\in \G$ with $L_S(g)=L_S(h).$ This shows $\min_{h\in \H}L_S(h)=\min_{g\in \G}L_S(g).$ So any computable ERM for $\G$ is also one for $\H$ and we are done by Fact \ref{fact scpac}.
\end{proof}

The above theorem immediately yields the following extension of Fact~\ref{fact scpac} and Fact~\ref{fact real rer}.
\begin{corollary}[Fundamental Theorem for proper realizable SCPAC]\label{cor rer}\

\noindent For any  hypothesis class $\H$ the following are equivalent:
\begin{enumerate}[itemindent=-1.3em]
    \item $\H$ is properly realizably SCPAC learnable.
    \item There is some RER class $\G\seq \H$  with $\vc(\G)<\infty$ and $\S_\H = \S_\G.$
    \item $\vc(\H)<\infty$ and there is a computable ERM $\A$ for $\H$ with $\S_\H\seq\dom(\A).$
\end{enumerate}
\end{corollary}

\begin{remark}[Non-strong learning]\label{rem cpac} The above proof essentially relied on Fact \ref{fact scpac}. Using an analogous result \cite[Proposition 13]{rose} for proper CPAC learnability, based on so-called \textit{asymptotic} ERMs, one can show an analogous version of Theorem \ref{thm rer scpac} for the non-strong case. Namely, a class $\H$ with $\vc(\H)<\infty$ is properly agnostically CPAC learnable if and only if there is some RER class $\G \seq \H$ for which there is a total computable asymptotic ERM $\A$ for $\G$ and a sequence $\eps_m\searrow 0$ such that $\min_{g\in \G}L_S(g)\leq \min_{h\in \H}L_S(h)+\eps_{|S|}$ for any sample $S.$ There is also a version in the realizable setting, where one needs the additional assumption on the domains. Details can be found in \cite[Corollary 3.41]{david}. \end{remark}

The proof of Fact \ref{fact evc rose} given in \cite[Theorem 11]{rose} via extension to a properly SCPAC learnable class essentially contains the following connection between CPAC learnability and RER classes, showing that finitely-supported agnostically CPAC learnable classes can be extended to RER classes. For details see \cite[Corollary 3.58]{david}.
\begin{corollary}\label{cor agn}
A hypothesis class $\H$ is agnostically CPAC learnable if and only if there is some RER class $\G\seq\Hfin$ with $\evc(\G)<\infty$ that satisfies $\S_\H \seq \S_\G.$ If
$\H\seq\Hfin,$ then it can even be ensured that $\H\seq \G$. 
\end{corollary}

To further highlight the connection between RER and CPAC we derive a partial converse of Fact \ref{fact real rer} by giving a sufficient condition under which CPAC learnability implies the RER property. It applies to the following type of hypothesis classes, which includes those of finite \textit{teaching dimension}, considered for instance in \cite{teaching,caro}. 

\begin{definition}\label{def uip}
We say that a hypothesis class $\H$ satisfies the \textbf{unique identification property} (UIP) if for every $h \in \H$ there are some $x_1,\hdots, x_k\in \X$ such that for any $h' \in \H\setminus \{h\}$ there is some $i\in [k]$ with $h(x_i)\neq h'(x_i).$
\end{definition}

\begin{lemma}\label{lem coincide}
Let $\H$ be realizably PAC learnable with some learner $\A$. Then for any $x_1,\hdots, x_k\in \X$ and $h\in \H$ there is some $g\in \im(\A)$ that coincides with $h$ on all $x_1,\hdots, x_k$.
\end{lemma}
\begin{proof}
Given some $h\in \H$ and any $x_1,\hdots, x_k\in \X$ consider the sample $S=((x_i,h(x_i)))_{i\in [k]}$ and the uniform distribution $\D_S=1/k\sum_{i=1}^k\delta_{(x_i,h(x_i))}$ over $S.$ It holds $L_{\D_S}(h')=L_S(h')$ for any hypothesis $h'.$ In particular, $L_{\D_S}(h)=0$ and thus $\D_S\in \Db_\H.$ So the realizable PAC learnability of $\H$ with learner $\A$ implies that for large enough $m$ we get
$$\prob{S'\sim \D_S^m}{L_{S}(\A(S'))\leq \frac{1}{|S|+1}}\geq 1/2>0.$$  
Observe that $L_S(\A(S'))\geq 1/|S|$ holds whenever $\A(S')$ does not coincide with $h$ on $x_1,\hdots, x_k.$ Thus, the positivity of the above probability implies that there must be at least one $S'\in \dom(\A)$ such that $g=\A(S')$ has the desired property.
\end{proof}

\begin{proposition}[Fundamental Theorem for UIP classes]\label{prop uip}
Let $\H$ be a hypothesis class satisfying the UIP. Then the following are equivalent:
\begin{enumerate}[itemindent=-1.3em]
\item $\H$ is properly realizably CPAC learnable.
\item $\H$ is properly realizably SCPAC learnable.
\item $\H$ is an RER class with $\vc(\H)<\infty$.
\item $\vc(\H)<\infty$ and there is a computable ERM $\A$ for $\H$ with $\S_\H\seq\dom(\A).$
\end{enumerate}
\end{proposition}
\begin{proof} 
(3)$\Rightarrow$(2) holds by Fact \ref{fact real rer}, (2)$\Leftrightarrow$(4) by Fact \ref{fact scpac} and (2)$\Rightarrow$(1) holds by definition. It remains to show (1)$\Rightarrow$(3). So let $\H$ be realizably CPAC learnable with some computable learner $\A\colon \S\dash \H.$
Due to Lemma~\ref{lem image}, it suffices to show $\H\seq \im(\A).$  Fix any $h\in \H$ and let $x_1,\hdots, x_k\in \X$ be such that no $h'\in \H\setminus \{h\}$ agrees with $h$ on all of those. By Lemma~\ref{lem coincide} there is some $g\in \im(\A)$ that agrees with $h$ on them. But as $\im(\A)\seq \H,$ this already implies $g=h.$ 
\end{proof}

\begin{remark} Besides the implication CPAC $\Rightarrow$ RER, the notable part in the above proposition is the equivalence of proper CPAC and SCPAC in the realizable setting. The same holds in the improper agnostic setting by Fact~\ref{fact evc rose}. However, it is demonstrated in \cite[Theorem 14]{rose} that the equivalence of proper CPAC and SCPAC learnability is not true in the agnostic case, even for RER classes satisfying the UIP. We do not know whether or not this holds for general classes in the realizable case. \end{remark}

\section{Nonuniform Learning}\label{sec nupac}
Fact \ref{fact real rer} showed that realizable CPAC learnability can be ensured for RER classes with finite VC-dimension. In this section we study how this is possible in an agnostic setting with arbitrary distributions, corresponding to no prior knowledge. Due to Theorem \ref{thm evc} below, we cannot achieve agnostic CPAC learnability for arbitrary RER classes of finite VC-dimension. Therefore, we consider the relaxed notion of nonuniform PAC learning where the sample size is allowed to depend on the learned hypothesis. This framework was introduced in \cite{benedek} and some computability aspects in it were already discussed in \cite{agar}.

\begin{definition}\label{def nucpac}
A hypothesis class $\H\seq \Y^\X$ is called \textbf{nonuniformly PAC learnable}, if there is a total learner $\A\colon \S\to \Y^\X$ and some sample size function $\mnu\colon \N^2\times \H\to \N$ such that for every distribution $\D\in \Db^*$ and all $a,b\geq 1$, $h\in \H$ and $m\geq \mnu(a,b,h)$ it holds
\begin{equation}\label{eq nucpac}
\prob{S\sim \D^m}{L_\D(\A(S))\leq L_\D(h)+1/a}\geq 1-1/b.
\end{equation}
\noindent $\H$ is called nonuniformly CPAC learnable if it nonuniformly PAC learnable with some computable learner in the sense of Definition \ref{def cpac}. Similarly, \textit{proper} nonuniform (C)PAC learnability refers to the learner $\A$ satisfying $\im(\A)\seq \H$.
\end{definition}

Our main result is the following notable consequence of the RER property.

\begin{theorem}\label{thm nucpac}
Every RER class is properly nonuniformly CPAC learnable.
\end{theorem}

The proof of this theorem is based on a standard learning paradigm for nonuniform PAC learning, called \textit{structural risk minimization}, see~\cite[Section 7]{sbd} and~\cite{benedek}. We give an adapted version fitted to our setting here:
Consider some decomposition of the hypothesis class $\H$ of the form $\H=\bigcup_{n\in \N_+}\H_n$ and define for each $h\in \H$
\begin{equation*}
    n_h\coloneqq \min\{n\colon h\in \H_n\}.
\end{equation*}
Also pick some increasing function $\omega\colon \N_+\to \N_+$ satisfying $\sum_{n=1}^\infty1/\omega(n)\leq 1$ and some error function  $\eps\colon \N_+^2\to (0,\infty)$ satisfying
\begin{align}
&\forall b\colon \eps(\cdot,b) \text{ monotonically converges to } 0,\label{eq mon1} \\
& \forall m\colon \eps(m, \cdot) \text{ is increasing}, \label{eq mon2}\\
&\forall m,b, n \forall \D\in \Db^*  \colon \prob{S\sim \D^m}{\sup\limits_{h\in \H_n}|L_\D(h)-L_S(h)|\leq \eps(m,b)}\geq 1-1/b.\label{eq uc assump}
\end{align}
A \textbf{structural risk minimizer}\label{def srm} (SRM) for $\H$ w.r.t.~$\eps$ is a function $\tilde{\A}\colon \N_+\times \S\to \H$ satisfying 
\begin{equation}\label{eq srm rule}
\tilde{\A}(b, S)\in \argmin\limits_{h\in \H}\big[L_S(h)+\eps(|S|,b\omega(n_h))\big]
\end{equation}
for any $b\geq 1$. A computable SRM is defined analogously to Definition \ref{def cpac}. SRMs can be used for nonuniform PAC learning by the following lemma, a broad adaption of~\cite[Theorem 7.4]{sbd}. For the interested reader we include a proof in Appendix \ref{app nupac}.

\begin{lemma}\label{lem nupac}
Fix some hypothesis class $\H=\bigcup_{n\in \N_+}\H_n$ and some $\eps$ as above. Moreover, let $\tilde{\A}$ be an SRM for $\H$ w.r.t.~$\eps$ and define $s(b)\coloneqq \min\{M\colon \eps(M, b\omega(b))\leq 1/2b\}$. Then for all $b\geq 1$ and $h\in \H$ with $n_h\leq b$, any $m\geq s(b)$ and any $\D\in \Db^*$ it holds
\begin{equation*}
\prob{S\sim \D^m}{L_\D(\tilde{\A}(b,S))\leq L_\D(h)+1/b}\geq 1-1/b.
\end{equation*}
Under these conditions, $\H$ is properly nonuniformly PAC learnable with the induced learner
\begin{equation*}
\A\colon  \S\to \H, \A(S)\coloneqq \tilde{\A}(t(|S|),S),
\end{equation*}
where $t(m)\coloneqq \max\{b\colon s(b)\leq m\}$, respectively $t(m)\coloneqq 1$ if there is no such $b$.
\end{lemma}

This formulation of structural risk minimization allows us to obtain a computable learner from a computable SRM, given computability of $s$ and $t$. The idea is that, if $\eps$ is unbounded in the second argument, it suffices to compare finitely many hypotheses to minimize the desired sum and this can be done in a computable way, given that $\H$ is RER and $\eps$ is simple enough.\\

\begin{proof}[Proof of Theorem \ref{thm nucpac}]
Let $(h_n)_{n\in \N}$ be a computable listing of an RER class $\H$. We apply structural risk minimization to the decomposition $\H=\bigcup_{n\in \N}\{h_n\}$ with weighting $\omega(n)=2n^2$. Consider the function
$$\eps\colon \N_+^2\to (0,\infty),\ \eps(m,b)\coloneqq \sqrt{b/2m}.$$
The conditions \eqref{eq mon1} and \eqref{eq mon2} are obvious. Condition \eqref{eq uc assump} follows from Hoeffding's Inequality, which states that for any $n\in \N$, $m\geq 1$, $\delta\in (0,1)$ and any distribution $\D\in \Db^*$ it holds
$$\prob{S\sim \D^m}{|L_\D(h_n)-L_S(h_n)|\leq \delta}\geq 1-2e^{-2m\delta^2}.$$
Applying this to $\delta=\sqrt{\frac{\ln(2b)}{2m}}\leq \eps(m,b)$ yields the desired condition.

It remains to construct a computable SRM for $\H$ w.r.t.~this function $\eps$. 
Once we have achieved this we are done, since our choice of $\eps$ gives computability of the functions $s(b)$ and $t(m),$ implying computability of the induced nonuniform PAC learner from Lemma \ref{lem nupac}. We describe how a computable SRM works:
Given any $b\geq 1$ and sample $S\in \S$, first determine the size $m$ of $S$. The goal is to find a hypothesis 
\begin{equation}\label{eq srm}
\tilde{\A}(b,S)\in \argmin\limits_{h\in \H}\big[L_S(h)+\eps(m,2bn_h^2)\big].
\end{equation}
This is equivalent to minimizing the function 
$$F(n)\coloneqq E_S(h_n)+m\eps(m,2bn^2)= E_S(h_n)+n\sqrt{mb},$$
where $E_S(h)$ denotes the number of errors $h$ makes on the sample $S$.
Note that we can computably compare values of $F$. Furthermore, we can compute some $N$ with $\min_{n\in \N}F(n)=\min_{n\leq N}F(n)$. 
Indeed, this is satisfied as soon as $N\sqrt{mb}\geq m+\sqrt{mb},$
since for any $k>N$
\begin{align*}
F(k)> N\sqrt{mb}\geq m+\sqrt{mb}\geq F(1)\geq \mini{n\leq N}F(n).
\end{align*}
Altogether we can minimize $F$ by computing such a value $N$, comparing the finitely many values $F(0),\hdots, F(N)$ and returning some minimizing $n^*\leq N$.
\end{proof}

Notably, the learner constructed in the above proof only requires polynomial sample sizes in $a,b$ and $n_h$. So in the previous terminology nonuniform CPAC learnability can even be achieved in a strong sense. Also, since proper implies improper learnability of subclasses we immediately get the following result.
\begin{corollary}\label{cor cont}
Every subclass of an RER class is nonuniformly CPAC learnable.
\end{corollary}

Since the above results do not need any assumptions on the (effective) VC-dimension, they emphasize that nonuniform CPAC learning is a relatively weak learnability notion. It is, however, not trivial, as \cite[Theorem 18]{agar} shows that the class of all computable hypotheses is not nonuniformly CPAC learnable. Still, the above corollary indicates that counterexamples to nonuniform CPAC learnability must be very large classes, as they cannot be contained in any RER classes such as $\Hfin$. A full converse of Theorem \ref{thm nucpac} can of course not be true, since nonuniform CPAC learnability is a relaxation of agnostic CPAC learnability and there are non-RER classes satisfying this notion \cite[Example 3]{sterk}. Instead, we get the following partial converse in the style of our results from the previous section.

\begin{corollary}\label{cor nu}
Let $\H$ be properly nonuniformly CPAC learnable. Then there exists some RER class $\G\seq\H$ with $\S_\G=\S_\H$.
\end{corollary}
\begin{proof}
Let $\H$ be nonuniformly CPAC learnable with some learner $\A\colon \S\to \H$. Consider the RER class $\G\coloneqq \im(\A)$ (Lemma \ref{lem image}) with $\G\seq\H$ and $\S_\G\seq \S_\H$. Fix any $S\in \S_\H$ and write $S=((x_i,h(x_i)))_{i\in [k]}$ for some $x_1,\hdots, x_k\in \X$ and $h\in \H$. Note that the nonuniform PAC learnability of $\H$ with the learner $\A$ implies PAC learnability of the singleton class $\{h\}$ with the learner $\A$. Hence, Lemma \ref{lem coincide} yields some $g\in \G$ that coincides with $h$ on $x_1,\hdots, x_k$, so $S\in \S_\G$.
\end{proof}

\noindent The above necessary condition is not sufficient. As a counterexample take $\G=\Hfin$ and $\H$ as the class of all computable hypotheses. 

\section{The Effective VC-Dimension}\label{sec evc} 
In this section we focus on the relationship between the VC-dimension and its effective counterpart. It is immediate from the definitions that for any hypothesis classes $\H$ it holds $\vc(\H)\leq \evc(\H)$ and $\evc(\G)\leq \evc(\H)$ for any subclass $\G\seq \H.$  In \cite[Theorem 15]{rose} a hypothesis class $\H$ with $\vc(\H)=1$ and $\evc(\H)=\infty$ was constructed. Building up on this construction, we can show that in general there is no connection between $\vc(\H)$ and $\evc(\H)$ other than $\vc(\H)\leq \evc(\H).$ We only sketch the construction here, details can be found in Appendix \ref{app evc}.

\begin{theorem}\label{thm evc}
For every $1\leq k\leq \ell\leq \infty$ there is some RER class $\H^{k,\ell}\seq \hfin$ such that $\vc(\H^{k,\ell})=k$ and $\evc(\H^{k,\ell})=\ell.$
\end{theorem}
\begin{proof}(Sketch).
The construction of $\H^{k,\ell}$ is analogous to the one from~\cite[Theorem 15]{rose}, which is defined via an enumeration of Turing machines and natural numbers $((T_j, k_j))_{j}$. By bounding $k_j\leq \ell-k+1$ we can control the effective VC-dimension to be $\ell-k+1$, while maintaining VC-dimension $1$. Both dimensions are then increased by $k-1$ by shifting all supports to start after $k$ and then using a class sum with the class of all hypotheses with support $\seq [k-1]$, which has VC-dimension $k-1$.
\end{proof}

\begin{remark}[CPAC bounds]\label{rem bound}
The above result shows that relying on the VC-dimension in quantitative PAC bounds can be misleading. Classical PAC bounds, such as \cite[Theorem 6.8]{sbd}, assert that the required sample size satisfies a bound of the form $m_\H(a,b)\in \mathcal{O}\big(a^2(\vc(\H)+\ln(b)\big).$ However, a careful inspection of the underlying proofs reveals that such bounds usually rely on the existence of an ERM learner for $\H.$ In the absence of a computable ERM for $\H,$ the above bound does not necessarily apply. In contrast, if $\evc(\H)$ is finite, then the proof of \cite[Theorem 11]{rose} shows that there exists an extension $\H' \supseteq \H$ such that $\vc(\H') \leq \evc(\H)+1$ and for which a computable ERM learner exists. Consequently, within the CPAC framework, the appropriate quantitative bound is $\mathcal{O}\big(a^2(\evc(\H)+\ln(b)\big),$ which may be much larger than the classical bound in the light of our theorem. It therefore becomes an important question to identify conditions under which equality of $\vc(\H)$ and $\evc(\H)$ can be ensured. 
\end{remark}

Equality of the two dimensions can be ensured under the strongest possible learnability assumptions. This is however not a real restriction from a practical point of view, as these conditions are almost always satisfied in practice, as discussed in \cite{hta}. Note that the converse implication is not true as \cite[Theorem 14]{rose} demonstrates.

\begin{theorem}\label{prop equal}
If there exists a total computable ERM for $\H$, then $\evc(\H)=\vc(\H).$ In particular, this holds if $\H$ is properly agnostically SCPAC learnable.
\end{theorem}
\begin{proof}
The ``in particular''-statement follows from the first claim and Fact \ref{fact scpac}.
Suppose that $\vc(\H)$ is finite, as otherwise $\evc(\H)=\infty$ holds trivially. Let $\A\colon \S\to \H$ be a computable ERM for $\H$ and $k\coloneqq \vc(\H).$
It suffices to describe a computable $k$-witness. 
Given any $x_0,\hdots, x_k\in \X$, first check if they are pairwise distinct. If not, output $0$. Otherwise, for each tuple $y=(y_0,\hdots, y_k)\in \Y^{k+1}$ run $\A$ on the sample $S_{y}\coloneqq ((x_0,y_0),\hdots, (x_k,y_k)).$
After each run, check if 
$L_{S_{y}}(\A(S_{y}))>0$ and output the first $y$ satisfying this. 
Since $\A$ is an ERM for $\H$, we have that $L_{S_{y}}(\A(S_{y}))>0$ if and only if $(h(x_0),\hdots, h(x_k))\neq y$ for all $h\in \H.$
Due to $\vc(\H)=k$ there exists at least one $y$ satisfying the latter, so the algorithm terminates.
\end{proof}

\begin{remark}\label{rem examples}
Theorem \ref{thm evc} highlights that being RER is not a sufficient condition for finiteness of the effective VC-dimension and therefore agnostic CPAC learnability. Even more, it establishes a whole family of new examples at once. Table \ref{table hkl} illustrates the learnability properties of $\H^{k,\ell}$ for different choices of $k$ and $\ell$, which follow from the facts in Section \ref{sec prelims} and Theorem \ref{prop equal}.

\begin{table}[ht]
\centering
\begin{tabular}{c|c|c|c|c|c}
 &
  PAC &
  \begin{tabular}[c]{@{}c@{}}realizable\\ CPAC\end{tabular} &
  \begin{tabular}[c]{@{}c@{}}agnostic\\ CPAC\end{tabular} &
  \begin{tabular}[c]{@{}c@{}}proper\\ agnostic\\ CPAC\end{tabular} &
  \begin{tabular}[c]{@{}c@{}}proper\\ agnostic\\ SCPAC\end{tabular} \\ \hline
$k=\ell<\infty$ & \yes & \yes & \yes & \yes & \yes \\ \hline
$k<\ell<\infty$ & \yes & \yes & \yes & \?    & \no  \\ \hline
$k<\ell=\infty$ & \yes & \yes & \no  & \no  & \no  \\ \hline
$k=\ell=\infty$ & \no  & \no  & \no  & \no  & \no 
\end{tabular}
\caption{Learnability of $\H^{k,\ell}$ for various choices of $k$ and $\ell$.}\label{table hkl}
\end{table}
\end{remark}

\section{Summary and Open Questions}\label{sec sum}
In this paper we studied the relationship between RER classes, computable PAC learning and the effective VC-dimension in an attempt to further recover the Fundamental Theorem of Statistical Learning in the computable framework. The implications between various notions of CPAC learnability and the RER property are illustrated in Figure~\ref{fig diag2}. All arrows represent strict implications with the conditions on the arrows being sufficient for the implications to hold. The colored arrows indicate our contributions.

\begin{figure}[ht]
    \centering
    \begin{tikzcd}[column sep=scriptsize, row sep=2em]
	{\text{agnostic CPAC}} && {\text{realizable CPAC}} \\
	& {\text{RER}} \\
	{\text{nonuniform CPAC}} && {\text{proper realizable SCPAC}}
	\arrow[from=1-1, to=1-3]
	\arrow[from=1-1, to=3-1]
	\arrow[curve={height=-8pt}, from=3-3, to=1-3]
    \arrow["{\text{RER}}", curve={height=-8pt}, from=1-3, to=3-3]
	\arrow["{\evc<\infty}"{description}, from=2-2, to=1-1]
	\arrow["{\vc<\infty}"{description}, from=2-2, to=1-3]
	\arrow[from=2-2, to=3-1, blue]
	\arrow["{\text{UIP}}"{description}, tail reversed, no head, from=2-2, to=3-3, blue]
\end{tikzcd}
\captionsetup{width=\textwidth, skip=5pt}
    \caption{Relationship between notions of CPAC learnability and the RER property.}
    \label{fig diag2}
\end{figure}

Regarding the relationship between CPAC learning and the RER property, we observed that several notions of CPAC learnability can also be characterized via existence of certain RER classes realizing the same samples, summarized in Table~\ref{table hg}. These results can be interpreted in the way that it is not a real restriction to only work with RER classes, since \emph{any} class satisfying one of these learnability notions can also be represented by some RER class satisfying the same notion. In other words, whenever we learn a hypothesis class $\H$ using a computable learner then we actually learn an RER class $\G$ that is similar enough to $\H$ for the learning to translate to $\H$. 
We also noted that the effective VC-dimension, which characterizes agnostic CPAC learnability, can take arbitrary values larger than the standard VC-dimension. However, equality of $\vc$ and $\evc$ can be guaranteed under learnability assumptions, which are usually satisfied in practice. Thus the classical PAC bounds for sample complexity still apply for practically relevant classes.

In the published version of this paper we asked two open questions. During the publication process, however, we became aware that both questions can be answered based on examples provided in~\cite{hta}. We discuss these examples in Appendix~\ref{sec answer}. Based on these observations, we ask a modified version of our original question that is inspired by the lack of a characterization of realizable CPAC learnability in the style of the Fundamental Theorem.

\begin{question}\label{qu}
Does for every realizably CPAC learnable class $\H$ exist some RER class $\G$ with $\vc(\G)<\infty$ and $\S_\H\seq \S_\G$?
\end{question}
\stepcounter{theorem}




\ \newline
\noindent \textit{Acknowledgments.} The first author would like to thank Salma Kuhlmann for her invaluable supervision in the making of his master's thesis. Both authors thank Matthias C.~Caro and Laura Wirth for helpful discussions that led to a better understanding of the subject and relevant literature. The first author was partially funded by the Austrian Science Fund (FWF) 10.55776/COE12. The second author started this line of research within the research project \textit{Fundamentale Grenzen von Lernprozessen in künstlichen neuronalen Netzen}, funded by Vector Stiftung as part of the program \textit{MINT-In\-no\-va\-tio\-nen 2022}. Both authors received partial project funding from the Network Platform \textit{Connecting Statistical Logic, Dynamical Systems and Optimization} of University of Konstanz.}

\bibliographystyle{alphaurl}
\bibliography{refs.bib}

\appendix


\section{Proof of Lemma \ref{lem nupac}}\label{app nupac}
We restate and prove Lemma \ref{lem nupac} by adapting some ideas from \cite[Section 7]{sbd} and \cite{benedek}.

\begin{lemman}
Fix some hypothesis class $\H=\bigcup_{n\in \N_+}\H_n$ and some $\eps$ as above. Moreover, let $\tilde{\A}$ be an SRM for $\H$ w.r.t.~$\eps$ and define $s(b)\coloneqq \min\{M\colon \eps(M, b\omega(b))\leq 1/2b\}$. Then for all $b\geq 1$ and $h\in \H$ with $n_h\leq b$, any $m\geq s(b)$ and any $\D\in \Db^*$ it holds
\begin{equation}\label{eq ab}
\prob{S\sim \D^m}{L_\D(\tilde{\A}(b,S))\leq L_\D(h)+1/b}\geq 1-1/b.
\end{equation}
Under these conditions, $\H$ is properly nonuniformly PAC learnable with the induced learner
\begin{equation}\label{eq A}
\A\colon  \S\to \H, \A(S)\coloneqq \tilde{\A}(t(|S|),S),
\end{equation}
where $t(m)\coloneqq \max\{b\colon s(b)\leq m\}$, respectively ~$t(m)\coloneqq 1$ if there is no such $b$.
\end{lemman}

\begin{proof} We first want to show the inequality~\eqref{eq ab}. Observe that $s$ is well-defined and monotonically increasing by the conditions on $\eps.$
Fix any $b\geq 1$ and any $m\geq s(b)$ and $\D\in \Db^*$. It holds
\begin{align*}
&\prob{S\sim \D^m}{\exists n\in \N\colon  \sup\limits_{h\in \H_n}|L_\D(h)-L_S(h)|>\eps(m,b \omega(n))}\\\leq &\summe{n=1}{\infty}\prob{S\sim \D^m}{\sup\limits_{h\in \H_n}|L_\D(h)-L_S(h)|>\eps(m,b \omega(n))}
\overset{\eqref{eq uc assump}}{\leq} \summe{n=1}{\infty}\frac{1}{bw(n)}\leq 1/b,
\end{align*} 
which is equivalent to
\begin{equation}\label{eq new S}
\prob{S\sim \D^m}{\forall n\in \N\colon  \sup\limits_{h\in \H_n}|L_\D(h)-L_S(h)|\leq \eps(m,b \omega(n))}
\geq 1-1/b.
\end{equation}
Now consider any $h\in \H$ with $n_h\leq b$.
The monotonicity assumptions on $\eps$ imply 
\begin{equation}\label{eq b}
    \eps(m,b\omega(n_h))\leq\eps_{b}(m,b\omega(b))\leq 1/2b.
\end{equation} Set $a\coloneqq \tilde{\A}(b,S).$ By the definition of SRMs any sample $S$ of size $m$ satisfying the inner condition in \eqref{eq new S} also satisfies
\begin{equation}\label{eq some S}
\begin{split}
L_\D(a)&\overset{\eqref{eq new S}}{\leq}  L_S(a)+\eps(m,b\omega(n_a))\\
&\overset{\eqref{eq srm rule}}{=}\mini{h'\in \H}\big[L_S(h')+\eps(m,b\omega(n_{h'})\big]\\
&\leq L_S(h)+\eps(m,b\omega(n_h)\\
&\overset{\eqref{eq b}}{\leq}  L_S(h)+1/2b\\
&\overset{\eqref{eq new S}}{\leq} L_\D(h) + \eps(m,b\omega(n_h))+ 1/2b\\
&\overset{\eqref{eq b}}{\leq} L_\D(h)+1/b.
\end{split}
\end{equation}
Together with \eqref{eq new S} this establishes \eqref{eq ab}.

Now we want to verify that the learner $\A$ defined in \eqref{eq A} is actually a nonuniform PAC learner for $\H$.
Due to our monotonicity assumptions the function $s$ is increasing and unbounded, so $t$ is well-defined. For any $(a,b,h)\in \N_+^2\times \H$ we pick 
\begin{equation}\label{eq mnus}
\mnu(a,b,h)\coloneqq s\big(\max\{a,b,n_h\}\big).
\end{equation}
Now fix any $a,b\geq 1,$ $h\in \H,$ $\D\in \Db^*$ and $m\geq \mnu(a,b,h).$ Then there is some $b'$ with $s(b')\leq m,$ namely $b'=\max\{a,b,n_h\}.$ So by definition $t(m)$ is the largest such $b'.$ This implies $t(m)\geq \max\{a,b, n_h\},$ and $m\geq s(t(m)).$
Hence \eqref{eq ab} yields
\begin{align*}
&\prob{S\sim \D^m}{L_\D(\A(S))\leq L_\D(h)+1/a}\\
\geq  &\prob{S\sim \D^m}{L_\D(\tilde{\A}(t(m),S))\leq L_\D(h)+1/t(m)}\\
\geq  &1-1/t(m)\geq 1-1/b.
\end{align*}
\end{proof}

\section{Proof of Theorem \ref{thm evc}}\label{app evc}
We generalize the construction from \cite[Theorem 15]{rose} to arbitrary values. Above we defined the VC-dimension via $k$-witnesses, but here we use the following standard definition of the VC-dimension.

\begin{definition}
A set $X\seq \X$ is said to be shattered by a hypothesis class $\H$ if for any function $f\colon X\to \Y$ there is some $h\in \H$ with $h\res_X =f.$
\end{definition}

\begin{lemma}\label{lem vc}
$\vc(\H)$ equals the size of the largest set $X\seq \X$ that is shattered by $\H.$
\end{lemma}
\begin{proof}
This follows from observing that there exists a $k$-witness for $\H$ if and only if $\H$ does not shatter any set $X\seq \X$ with at least $k+1$ elements.
\end{proof}

\noindent Our construction uses a shifting trick taking the direct sum of two hypothesis classes. Under suitable conditions, the VC-dimension of such a sum equals the sum of the individual VC-dimensions.

\begin{lemma}\label{lem sum}
Let $\G$ and $\H$ be hypothesis classes both containing the zero hypothesis and with disjoint supports, i.e.,~$\supp(g)\cap \supp(h)=\emptyset$ for all $g\in \G, h\in \H.$  Then the \textbf{direct sum} $\G\oplus \H\coloneqq \{g+h\colon g\in \G, h\in \H\}$ is a hypothesis class satisfying $\vc(\G\oplus\H)=\vc(\G)+\vc(\H).$
\end{lemma}
\begin{proof}
Write $G\coloneqq \bigcup_{g\in \G}\supp(g)$ and $H\coloneqq \bigcup_{h\in \H}\supp(h)$ and $\F\coloneqq \G\oplus\H$. Without loss of generality assume $|\F|>1,$ otherwise the statement is trivial. Let $X\seq \X$ be shattered by $\F$. Partition $X$ into three disjoint sets
$X=(X\cap G)\cup (X\cap H)\cup (X\setminus (G\cup H)).$
Since $X$ is shattered by $\F$, its subsets $X\cap G$, $X\cap H$ and $X\setminus (G\cup H)$ are also shattered by $\F$. By the choice of $G$ and $H$, any $f\in \F$ is $0$ on $X\setminus (G\cup H),$ so this set must be empty. 
From $G\cap H=\emptyset$ it follows that any $g\in \G$ is $0$ on $X\cap H.$ Hence, $X\cap H$ must be shattered by $\H$ (because also $0\in \H$). Analogously, $\G$ shatters $X\cap G$. This shows $|X|=|X\cap G|+|X\cap H|\leq \vc(\G)+\vc(\H)$ and thus $\vc(\G\oplus\H)\leq \vc(\G)+\vc(\H)$. For `$\geq$', let $X_1\seq \X$ be shattered by $\G$ and $X_2\seq \X$ be shattered by $\H.$ Then $X_1\seq G$ and $X_2\seq H$, so $X_1$ and $X_2$ must be disjoint sets.
Hence, it suffices to show that $X_1\cup X_2$ is shattered by $\F.$  For any $f\colon X\to \Y$ there are some $g\in \G$ and $h\in \H$ with $g\res_{X_1}=f\res_{X_1}$ and $h\res_{X_2}=f\res_{X_2}.$ Since $h$ is $0$ on $X_1$ and $g$ is $0$ on $X_2$ it follows $f=(g+h)\res_{X_1\cup X_2}.$
\end{proof}


\noindent Now we we can prove the theorem. 
\begin{theoremn}
For every $1\leq k\leq \ell\leq \infty$ there is some RER class $\H^{k,\ell}\seq \hfin$ with $\vc(\H^{k,\ell})=k$ and $\evc(\H^{k,\ell})=\ell.$
\end{theoremn}

\begin{proof}
In the case $k=\ell,$ we choose $\H^{k,\ell}$ to be the RER class of all hypotheses with support size at most $k,$ then we are done by Example~\ref{ex rer}. So we assume $k<\ell$ in the following. Also assume that $k$ is even, the odd case works analogously.\vspace{1,5mm}

\noindent \textbf{I.~Preliminary work:} For a Turing machine $T$ let $\code(T)\in \N$ be the goedelization of $T$ (see \cite[p.\,10]{soare}). Consider an enumeration $((T_j,k_j))_{j\in \N}$ of all pairs of Turing machines and numbers from $[\ell-k+1]$ for which the maps $j\mapsto \code(T_j)$ and $j\mapsto k_j$ are computable. This exists because recursive enumerability is preserved under cartesian products. Let $(I^j)_{j\in \N}$ be a computable enumeration of tuples of even numbers $\geq k$ such that $I^j$ has length $k_j$ and every even number $\geq k$ appears in exactly one tuple $I^j$. 
Now define $E\coloneqq \big\{e\in \N \colon \exists o^e\in \Y^{k_e}\colon T_e(I^e)=o^e\big\},$ the set of all indices $e\in \N$ for which $T_e$ halts on the input $I^e\in \N^{k_e}$ with some binary output $o^e$ of length $k_e$. Note that $E$ is infinite and $o^e$ is well-defined for $e\in E.$
Lastly, for $e\in E$ let $s_e$ be the number of steps after which $T_e$ halted on the input $I^e$ and define $u_e \coloneqq  2\cdot 3^e\cdot 5^{s_e}+k+1.$\vspace{1,5mm}


\noindent \textbf{II.~Construction of $\H^{k,\ell}$:} 
Consider the hypothesis class 
\begin{equation}\label{eq he}
\H_E\coloneqq \{h_e\colon e\in E\}\cup \{0\}, \text{ where }
h_{e}(x)\coloneqq \begin{cases}
1, &\text{if } x=u_e\\
o^e_i, & \text{if } x=I^e_i\\
0, & \text{else.}
\end{cases}
\end{equation}
Note that the support of $h_e$ contains only entries of $I^e$ and the single odd number $u_e$. So the hypotheses in $\H_E$ have disjoint supports, which implies $\vc(\H_E)=1.$
Next, consider the class $\G\coloneqq \{g\in \Y^\N\colon \supp(g)\seq [k-1]\},$
which satisfies $\vc(\G)=k-1.$ As $\supp(h_e)$ does not contain any elements smaller than $k,$ the supports of hypotheses from $\G$ and $\H_E$ are always disjoint. Lemma \ref{lem sum} now yields that their direct sum
$\H^{k,\ell}\coloneqq \G\oplus\H_E$ has VC-dimension exactly $k$.\vspace{1,5mm}

\noindent \textbf{III.~Decidability of $\H^{k,\ell}$:} We describe how membership in $\H^{k,\ell}$ can be decided from the list representation of $\Hfin$. This already gives that $\H^{k,\ell}$ is RER, as $\Hfin$ is an RER class itself.

Given any $h\in \Hfin,$ first check if there is at most one odd number $x> k$ in $\supp(h)$. Return NO if more than one $x$ is found. If there is none, check for $\supp(h)\seq [k-1]$ and return YES, if this is true and NO otherwise.  
If there is exactly one such odd number $x$, try to compute some $e,s\in \N$ with $x=2\cdot 3^e\cdot 5^s+k+1.$
If no such $e$ and $s$ exist, output NO. Otherwise, compute $\code(T_e)$, $k_e$ and $I^e$ and run (at most) $s$ steps of $T_e$ on the input $I^e$. If $T_e$ halts after exactly $s$ steps, check if the output is some tuple $o^e\in \Y^{k_e}.$ This is true if and only if $e\in E,$ $s=s_e$ and $x=u_e$. Now verify $h(I_i^e)=o^e_i$ for all $i=1,\hdots, k_e.$ Lastly, check if all elements of $\supp(h),$ except $x$ and the ones from $I^e,$ are elements of $[k-1].$ If all of this is true, then it must hold $h=g+h_e$ for some $g\in \G$, thus return YES. In any other case output NO.\vspace{1,5mm}

\noindent $\textbf{IV.~eVCdim(}\H^{k,\ell}\textbf{)}\leq \ell\colon $ We describe a computable $\ell$-witness for $\H^{k,\ell}$ in the case $\ell<\infty$. So we have to find a computable function $w\colon \N^{\ell+1}\to \Y^{\ell+1}$ such that for all $h\in \H^{k,\ell}$ and distinct $x_0,\hdots, x_\ell\in \X$ it holds
\begin{equation}\label{eq dagger}
w(x_0,\hdots, x_\ell)\neq (h(x_0),\hdots, h(x_\ell)). 
\end{equation}
Given any input $(x_0,\hdots, x_\ell)\in \X^{\ell+1}$, first check if all entries are pair-wise distinct. If not, $w$ outputs $0$. Otherwise, we use the decision procedure described in step III to check whether the hypothesis $f$ with $\supp(f)=\{x_0,\hdots, x_\ell\}$ lies in $\H^{k,\ell}$. In the case $f\notin \H^{k,\ell}$, $w$ simply returns $(1,\hdots, 1).$ Then the desired condition \eqref{eq dagger} holds for any $h\in \H^{k,\ell}$ due to the fact $|\supp(h)|\leq \ell+1.$ 

Now consider the case $f\in \H^{k,\ell}.$ By the construction of $\H^{k,\ell}$  there must be some $e\in E$ such that $k_e=\ell-k+1,$ $o^e=(1,\hdots, 1)$ and $f=\one_{[k-1]}+h_e.$ In particular, $\{x_0,\hdots, x_\ell\}=[k-1]\cup \supp(h_e).$
Using the same procedure as in step III, we can compute this index $e\in E$ and the $i\leq \ell$ with $x_i=u_e$, which is the unique odd number $>k$ in $\supp(h_e).$ Now $w(x_0,\hdots, x_\ell)$ is defined as the tuple that is $1$ everywhere, except at the position $i,$ where we set it to $0$. With this choice condition \eqref{eq dagger} is satisfied for any $h\in \H^{k,\ell}$ with $h(x_i)=1.$ 
Now consider $h\in \H^{k,\ell}$ with $h(x_i)=0.$ Write $h=g+h_{e'}$ for some $e'\in E$ and $g\in \G.$ Because of $h_{e'}(x_i)=0,$ it must hold $e'\neq e.$ By assumption we have $|\supp(h_e)|=k_e+1\geq 2,$ so there is some $x_j\in \supp(h_e)\setminus \{x_i\}$. From $\supp(h_{e'})\cap \supp(h_e)=\emptyset,$ it follows $h(x_j)=0.$ So \eqref{eq dagger} holds for $h$ as $w(x_0,\hdots, x_\ell)$ is 1 at position $j$.\vspace{1,5mm}

\noindent $\textbf{V.~eVCdim(}\H^{k,\ell}\textbf{)}\geq \ell\colon $ 
It suffices to show that $\H^{k,\ell}$ has no computable $(\ell-1)$-witness. If $\ell=\infty$ replace $\ell$ by any number $\geq k$ in the following. 
Assume that there was a computable  $(\ell-1)$-witness $w\colon \X^\ell\to \Y^\ell$ for $\H^{k,\ell}$. Consider the computable function $\tilde{w}\colon \X^{\ell-k+1}\to \Y^{\ell-k+1}$ with
\begin{align*}
\tilde{w}(x_0,\hdots, x_{\ell-k})= \text{last $(\ell-k+1)$ entries of } w(1,\hdots, k-1, x_0,\hdots, x_{\ell-k}).
\end{align*}
By construction there is some $e\in E$ such that $T_e$ computes the function $\tilde{w}$ and $k_e=\ell-k+1$. Thus by our choices
\[\tilde{w}(I^e)=o^e=(h_e(I^e_1),\hdots, h_e(I^e_{k_e})).\]
Also note that there is 
$g\in \G$ with 
\[(g(1),\hdots, g(k-1))= \text{the first $(k-1)$ entries of } w(1,\hdots, k-1,I^e).\]
Altogether, the hypothesis $h\coloneqq g+h_e\in \H^{k,\ell}$ satisfies
\begin{align*}
w(1,\hdots, k-1,I^e)
=(h(1),\hdots, h(k-1),h(I^e_1),\hdots, h(I^e_{k_e})).
\end{align*}
This contradicts $w$ being an $(\ell-1)$-witness for $\H^{k,\ell}.$ \end{proof}

\section{Answers to Some Questions}\label{sec answer}
\noindent In this section we answer both questions from the published version of this paper, utilizing the construction from \cite[Theorem 5.1]{hta}.
Our first question was inspired by the lack of a characterization of realizable CPAC learning similar to our results from Section~\ref{chap rer}:

\renewcommand{\thetheorem}{31*}
\begin{question}
Can every realizably CPAC learnable class be extended to an RER class with finite VC-dimension?
\end{question}

The below counterexample shows the requirement of direct extendability is too strong, so the general answer to that question is no. Instead, in analogy to Corollary~\ref{cor agn}, a potential characterizing condition for realizable CPAC learnability, which also fits to our counterexample, could be the existence of some RER class $\G$ with $\vc(\G)<\infty$ and $\S_\H\seq \S_\G$. While this condition is easily seen to be sufficient, it is unclear whether it is also necessary, see Question \ref{qu}.

The second question we asked was related to the effective VC-dimension of classes containing uncomputable hypotheses:

\renewcommand{\thetheorem}{32*}
\begin{question}
Are there any CPAC learnable classes containing uncomputable hypotheses?
\end{question}

The answer to this question is a clear yes and in fact the same example as the one refuting the first question demonstrates this. 

\begin{example}
Similar to~\cite[Theorem 5.1]{hta}, we consider the class of real decision stumps $\H=\{h_x=\one_{(-\infty, x)}\colon x\in \R\}$ on the domain $\X=\Q$. The class is not contained in any RER class, since for every uncomputable real number $x$, the corresponding hypothesis $h_x$ is uncomputable~\cite{weihrauchcomp}, but RER classes only contain computable hypotheses.  However, $\H$ contains the RER class $\G=\{h_x\colon x\in \Q\}$, which has $\vc(\G)=1$ and satisfies $\S_\G=\S_\H$. Thus, $\H$ is realizably CPAC learnable by Corollary~\ref{cor rer}. Moreover, we can construct a computable ERM for $\G$ by ordering the domain points $x_1<\dots < x_m$ of a given sample and then test the $m+1$ decision stumps $h_{x_1},\dots, h_{x_m}, h_{x_m+1}$ to pick the one with minimal loss (cf.~\cite[Theorem 13]{agar}). Hence, both $\G$ and $\H$ are even properly agnostically SCPAC learnable by Theorem~\ref{thm rer scpac} and on top of that Theorem~\ref{prop equal} yields $\evc(\H)=\vc(\H)=1$.
\end{example}

\end{document}